# Analysis of lane-change conflict between cars and trucks at merging section using UAV video data

Yichen Lu[a], Kai Cheng[b], Yue Zhang[a], Xinqiang Chen[c] and Yajie Zou[a]*

[a] *Key Laboratory of Road and Traffic Engineering of Ministry of Education, Tongji University, Shanghai 201804, China;* [b]*China Railway Siyuan Survey and Design Group Co., Ltd., Wuhan 430063, China;* [c] *Institute of Logistics Science and Engineering, Shanghai Maritime University, Shanghai 201306, China*

Correspondence should be addressed to: yajiezou@hotmail.com

# Analysis of lane-change conflict between cars and trucks at merging section using UAV video data

The freeway on-ramp merging section is often identified as a crash-prone spot due to the high frequency of traffic conflicts. Very few traffic conflict analysis studies comprehensively consider different vehicle types at freeway merging section. Thus, the main objective of this study is to analyse conflicts between different vehicle types at freeway merging section. Field data are collected by Unmanned Aerial Vehicle (UAV) at merging areas in Shanghai, China. Vehicle extraction method is utilized to obtain vehicle trajectories. Time-to-collision (TTC) is utilized as the surrogate safety measure. TTC of car-car conflicts are the smallest while TTC of truck-truck conflicts are the largest. Traffic conflicts frequently occur at on-ramp and acceleration lane. Results show the spatial distribution of lane-change conflicts is significantly different between different vehicle types, suggesting that vehicle drivers should maintain safe distance especially car drivers. Besides, in order to decrease lane-change conflict at merging area, traffic management agencies are suggested to change dotted lie to solid lane at the beginning of acceleration lane.

Keywords: lane-change conflict; time-to-collision; merging section; UAV

## 1 Introduction

The merging section is a significant segment of the freeway due to frequent interaction between lane-changing vehicle and surrounding vehicles merging section (Li et al., 2016). Hence, it is important to analyse traffic conflict characteristics at these sections to improve traffic operation and safety. Many studies have evaluated of the safety performance at merging sections using crash data. However, this method has its own limitations. For example, crashes are unpredictable rare events and not all crashes are reported (Laureshyn et al., 2010; Oh et al., 2010; Svensson & Hyden, 2006). Consequently, in order to overcome these shortcomings, traffic conflict technique was proposed. This method can observe a large amount of data before crashes, with the statistical advantages of large sample and short cycles.

Traffic conflict technique is widely used for road safety analysis nowadays (L. Zheng et al., 2018; Lai Zheng & Sayed, 2019). In order to measure the temporal and spatial proximity of interactions between road users, numerous traffic conflict indicators have been developed. In previous studies, TTC and post encroachment time (PET) are two common traffic conflict indicators (A. Y. Chen et al., 2020; Meng & Qu, 2012; Qi et al., 2020; Lai Zheng & Sayed, 2019; L. Zheng et al., 2019). Chin et al. (1991) utilized the inverse of TTC to measure the severity of conflicts at merging sections in Singapore. Uno et al. (2002) utilized TTC and (Potential Index for Collision with Urgent Deceleration) PICUD to analyse rear-end conflict caused by lane-change vehicle at target lane in weaving section. Yang and Ozbay (2011) proposed a two-step method to estimate rear-end conflict risk of vehicles on freeway merging section. The first step focus on estimating the merging probability of a vehicle. Then, modified TTC was utilized as a surrogate safety measure. These two steps were combined and a conflict index was proposed. Li et al. (2016) defined the hourly composite risk indexes (HCRI) based on TTC to evaluate the traffic safety of freeway interchange merging sections. Gu et al. (2019) calculated the crash risk between the merging vehicle and surrounding vehicles by incorporating TTC and the estimated merging behaviour's model at interchange merging sections. Lai et al. (2020) utilized PET as the conflict indicator to investigate the safety performance at different merging sections. Although PET also sometimes utilized in traffic conflict analysis at merging sections, TTC can reflect more information than PET (Mahmud et al., 2017), this study consequently considers TTC as the traffic conflict indicator.

Additionally, collecting high accurate vehicle trajectory data is also important for traffic conflict study. Video surveillance has been used to obtain vehicle trajectory data with the development of the computer vision and video processing technique. One traditional vehicle trajectory data collection approach is the video surveillance using fixed cameras. However, this approach has some disadvantages (Gu et al., 2019). For example, videos collected by fixed cameras are always at a tilt angle, so vehicles may be occluded by neighboring vehicles. It is also difficult to detect vehicles far away from the fixed camera (X. Chen et al., 2021). Besides, video shooting range is restricted by the height of the fixed camera. In order to overcome the above shortcomings, the UAV has been utilized to collect vehicle trajectory data in some studies (Gu et al., 2019; Li et al., 2016). The UAV flies can not only cover a larger range, but also avoid occlusion between neighboring vehicles. Some open source dataset used UAV as the vehicle trajectory recording equipment, such as the inD Dataset (Bock et al., 2020), the highD dataset (Krajewski et al., 2018).

Cars and trucks have different operating characteristic, such as size and braking capability. Some studies examined the time headway between different types of vehicles. Ye and Zhang (2009) categorized four headway groups based on vehicle types. Results showed that car-car headway is the smallest while the truck-truck headway is the greatest. Truck-involved headway is larger because of the large volume and the poor braking ability of trucks. Zou et al. (2017) found that when speed is lower than 20 kph, vehicle type is an important influence on following headway. Previous studies reveal that vehicle type plays an important role for traffic safety analysis. Besides, some research focus on rear-end conflicts between different vehicle types. Meng and Weng (2011) proposed rear-end risk model to evaluate the rear-end crash risk at work zone activity area and merging section. Deceleration rate to avoid the crash (DRAC) was utilized to measure rear-end crash risk. Compared with cars, trucks have a higher probability involving rear-end accidents at work zone activity area. Weng et al. (2015) had the similar findings that the rear-end crash risk for merging vehicles is significantly affected by vehicle types. For merging vehicles in work zone merging areas, when the merging lead vehicle is a heavy vehicle, the rear-end crash risk is greatly increased.

However, few studies have evaluated the lane-change conflicts between different vehicle types at the merging section. Considering the size and kinematic difference between cars and trucks result in different driving behaviours, the objective of this study is to explore the characteristics of lane-change conflicts between cars and trucks at the merging section with high proportion of trucks. The UAV is utilized to collect data. Vehicle trajectory information is extracted through the proposed approach. Time-to-collision is utilized as the traffic conflict indicator in this study.

## 2 Methodology

### 2.1 Vehicle trajectory processing

Vehicle trajectory processing includes three steps (X. Chen et al., 2021). The first step is vehicle detection with ensemble detector. Then, the second step is vehicle tracking by KCF algorithm. The final step is trajectory denoising through WT. As Figure 1 shows the results of each step during vehicle trajectory processing.

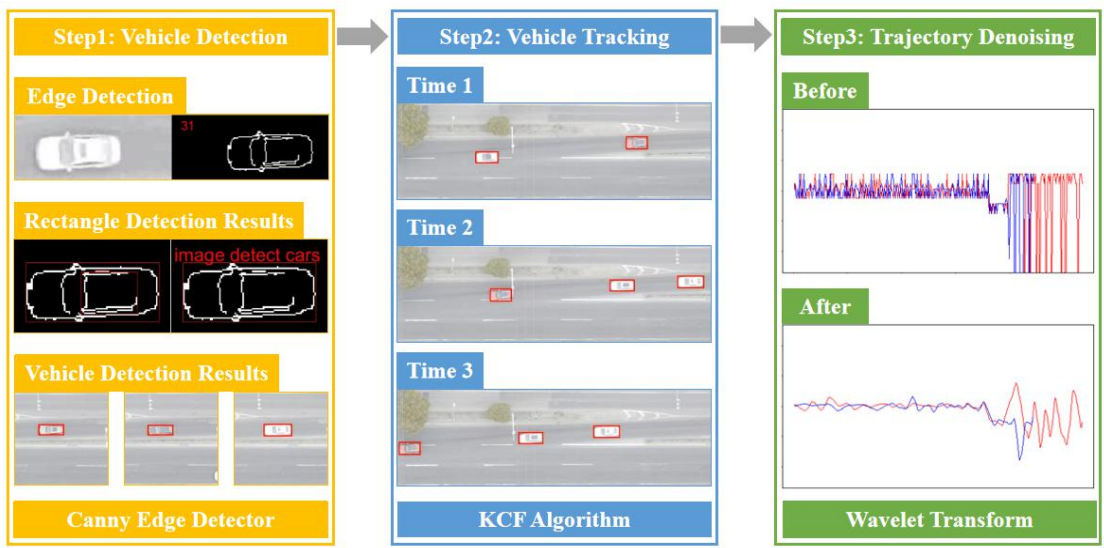

Figure 1. Vehicle trajectory processing

*2.1.1 Vehicle detection*

Before vehicle detection, the region of interest (ROI) is set on each lane at the start of the observed road segment. Vehicle detection actually involving three stages. At the first stage, an ensemble Canny based edge detector is employed to detect vehicles. When vehicle edge gradient or intensity is lower than threshold, the Canny edge detector may see these edge pixels as background and suppress them from vehicle edges. In order to obtain smooth and accurate edges, Morphology close operation is utilized. At the second stage, based on connectivity criterion, edges should be merged into rectangles. Each rectangle represents one vehicle. The final stage is data quality control, which focuses on removing obvious detection outliers by the ensemble detector. For example, some roadway pixel may be detected as vehicles, so length-width ratio threshold of vehicles should be set to avoid this situation. Some other thresholds also need to be set in order to remove outliers. Through the third stage, only correct results will be retained.

*2.1.2 Vehicle tracking*

The KCF algorithm is widely used in target tracking (Olfati-Saber, 2009; Tseng et al., 2021). Due to top-view recording angle, trees and other obstacles may sheltered vehicles. Those obstacles may be tracked as vehicles sometimes. However, KCF algorithm avoid the above situation, so that vehicle tracking results can be better. First of all, we should train a vehicle tracker. The process of training a vehicle tracker is a process of find the optimal solution. Based on the optimal solution, vehicles can be tracked by KCF algorithm in each frame easily. However, tracking results still need careful check in order to suppress outliers. For instance, one typical outlier is that some neighbour vehicles have similar tracking positions. These outliers always occur when length or width of a vehicle is larger than length or width of the ROI in each lane. Therefore, neighbour vehicles' position should be merged into one vehicle's position. Through data quality control, precise vehicle trajectories can be obtained.

*2.1.3 Trajectory denoising*

Although data quality control has been carried out in each previous step, there are still some small errors exist. For example, due to the vibration of tracking rectangle, vehicle trajectories may have irregular oscillation. Trajectory denoising with WT aims to eliminate these small errors, so that vehicle trajectories can be more accurate. Raw vehicle trajectory data is divided into two subsets through WT, one group is scaling subsets, and another group is wavelet subsets. The wavelet subsets include details and noises in raw vehicle trajectories. The vibration of tracking rectangle results in the irregular oscillation in raw vehicle trajectories. These irregular oscillation leads to high fluctuation margin of wavelet subsets. The high fluctuation margin can be seen as white noises. White noises must be smoothed out through setting up proper thresholds, so that non-noise wavelet vehicle trajectories can be obtained. Through trajectory denoising process, white noses are suppressed so that high accuracy vehicle trajectories can be obtained.

*2.2 Definition of lane-change conflicts*

Due to the characteristic of freeway, only lane-change conflicts and rear-end conflicts will occur. In our study, we only focus on lane-change conflicts. For each traffic conflict, when at least on vehicle is changing lanes, this traffic conflict is defined as a lane-change conflict. If no vehicle is changing lanes, this situation needs further discussion. When two cars are not at the same lane, even no vehicle is changing lanes, this situation also belongs to the lane-change conflict.

*2.3 Time-to-collision in lane-change conflict*

TTC was defined by (Hayward, 1972) as the time required for two vehicles to collide if two vehicles maintain their present speed and on the same path. TTC can be calculated by the following equation (Minderhoud & Bovy, 2001):

$$TTC_i = \frac{x_i - x_j - l_i}{v_i - v_j} \qquad \forall v_i > v_j \qquad (1)$$

where $x_i$ and $x_j$ is the position of vehicle $i$ and vehicle $j$, $v_i$ and $v_j$ is the speed of vehicle $i$ and vehicle $j$, and $l_i$ is the length of vehicle $i$.

However, this TTC only suitable for a one-dimension problem. In order to adapt to various conflict scenarios, some studies revise TTC into a two-dimension problem (Hou et al., 2014; Laureshyn et al., 2010; Ward et al., 2015). Similarly, TTC also considered as a two-dimension problem in our study.

The challenge of calculating TTC is the determination of the collision point. The actual situation of a vehicle collision is a point-to-point collision of two vehicles. As Figure 2 shows, these two points are defined as the collision points of these two vehicles. Each vehicle is treated as a rectangle during the calculation, and each vehicle has four sides and four corners.

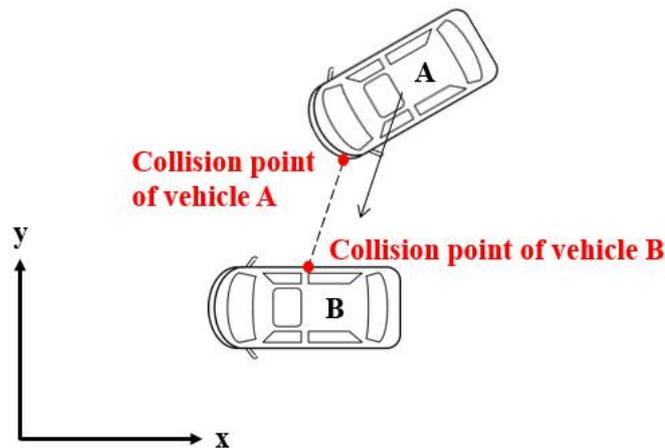

Figure 2. Collision points of two vehicles

When relative speed and relative angle of two vehicles are not be determined, collision points of the two vehicles will have different situations. However, for different situations, the common point is that if there is a conflict between two vehicles, the two collision points must include at least one vehicle's corner. According to this feature, TTC can be calculated by the following method. It should be noted that the coordinate is established during vehicle trajectory extraction, so the calculation of TTC is carried out in this coordinate system.

The length of vehicle A is $l_A$, the width of vehicle A is $w_A$, while the length of vehicle B is $l_B$, the width of vehicle B is $w_B$. For vehicle A, the coordinates of the four corners are $A_1(x_{A_1}, y_{A_1})$, $A_2(x_{A_2}, y_{A_2})$, $A_3(x_{A_3}, y_{A_3})$, $A_4(x_{A_4}, y_{A_4})$. For vehicle B, the coordinates of the four corners are $B_1(x_{B_1}, y_{B_1})$, $B_2(x_{B_2}, y_{B_2})$, $B_3(x_{B_3}, y_{B_3})$, $B_4(x_{B_4}, y_{B_4})$. The centre point of vehicle A is $(x_A, y_A)$, while the center point of vehicle B is $(x_B, y_B)$. The speed of vehicle A is $(v_{Ax}, v_{Ay})$, while the speed of vehicle B is $(v_{Bx}, v_{By})$. First of all, TTC should be calculated only when two vehicles are approaching. Vehicle B is seen as the stationary vehicle, and the relative speed of vehicle A is $(v_{Ax} - v_{Bx}, v_{Ay} - v_{By})$. As Figure 3 shows, the equation of a straight line $l_{A_i}$ passing through point $A_i$ and in the direction of relative speed is calculated as follows:

$$\frac{x - x_{A_i}}{v_{Ax} - v_{Bx}} = \frac{y - y_{A_i}}{v_{Ay} - v_{By}} \qquad (2)$$

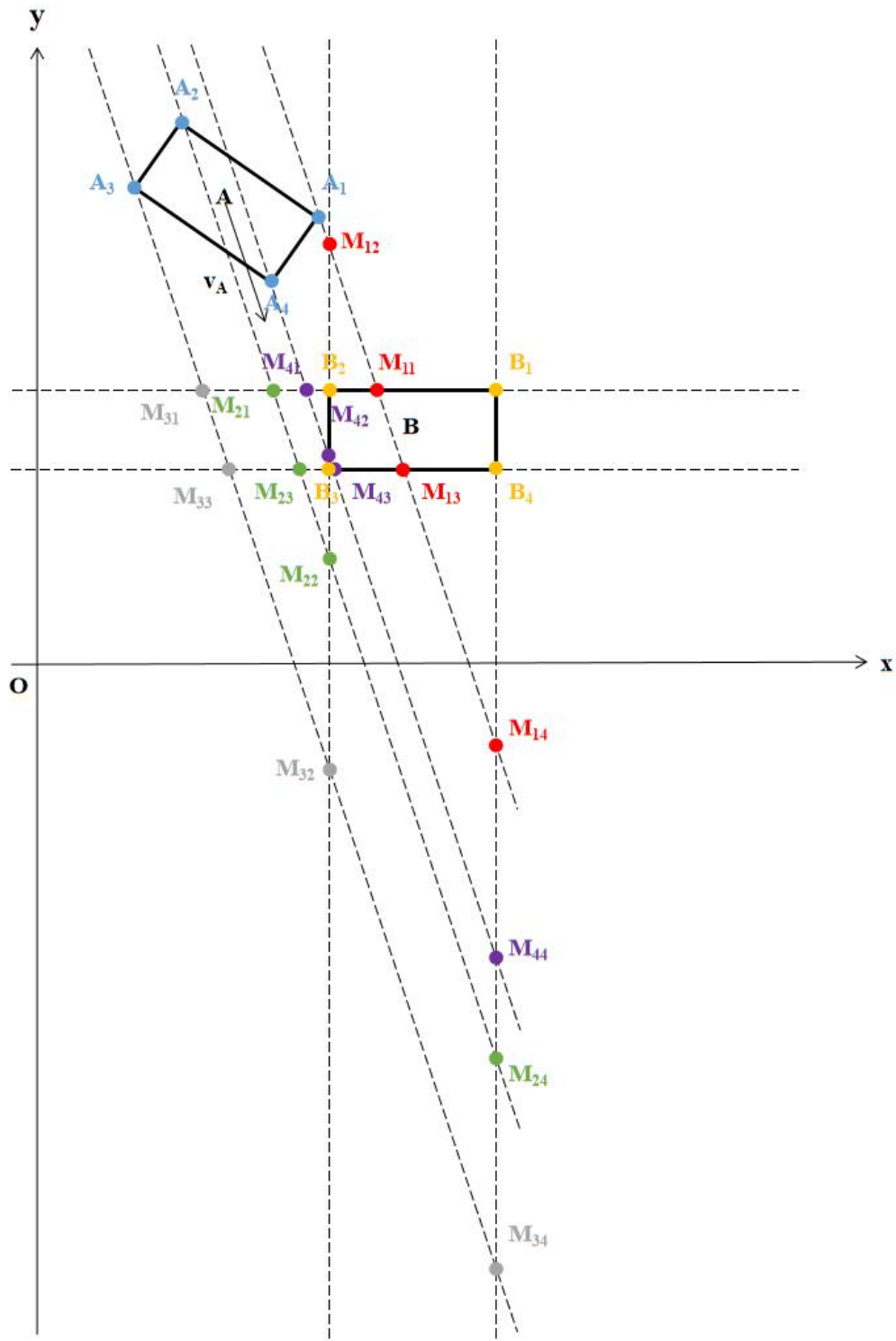

Figure 3. Calculation of traffic conflict between two vehicles

For each side of vehicle B, the equation of the straight line $l_{B_iB_j}$ passing through each side are easy to be obtained. For each $A_i$, intersection between $l_{A_i}$ and $l_{B_iB_j}$ can be calculated by the following four systems:

$$\begin{cases} \dfrac{x-x_{A_i}}{v_{Ax}-v_{Bx}} = \dfrac{y-y_{A_i}}{v_{Ay}-v_{By}} \\ \dfrac{x-x_{B_1}}{x_{B_1}-x_{B_2}} = \dfrac{y-y_{B_1}}{y_{B_1}-y_{B_2}} \end{cases} \text{and} \begin{cases} \dfrac{x-x_{A_i}}{v_{Ax}-v_{Bx}} = \dfrac{y-y_{A_i}}{v_{Ay}-v_{By}} \\ \dfrac{x-x_{B_2}}{x_{B_2}-x_{B_3}} = \dfrac{y-y_{B_2}}{y_{B_2}-y_{B_3}} \end{cases};$$

$$\begin{cases} \dfrac{x-x_{A_i}}{v_{Ax}-v_{Bx}} = \dfrac{y-y_{A_i}}{v_{Ay}-v_{By}} \\ \dfrac{x-x_{B_3}}{x_{B_3}-x_{B_4}} = \dfrac{y-y_{B_3}}{y_{B_3}-y_{B_4}} \end{cases} \text{and} \begin{cases} \dfrac{x-x_{A_i}}{v_{Ax}-v_{Bx}} = \dfrac{y-y_{A_i}}{v_{Ay}-v_{By}} \\ \dfrac{x-x_{B_4}}{x_{B_4}-x_{B_1}} = \dfrac{y-y_{B_4}}{y_{B_4}-y_{B_1}} \end{cases}.$$

(3)

Each system of linear equations have one solution, so four solutions are $M_{i1}(x_{M_{i1}}, y_{M_{i1}})$, $M_{i2}(x_{M_{i2}}, y_{M_{i2}})$, $M_{i3}(x_{M_{i3}}, y_{M_{i3}})$, $M_{i4}(x_{M_{i4}}, y_{M_{i4}})$. If one of system of linear equations has no solution, it means two straight lines are parallel, so there is no intersection. Each $A_i$ has four solution, so there are $4 \times 4$ solutions.

After that, the validness of the intersection also needs to be considered. The solution is valid only when the intersection is on a certain side of vehicle B. For the valid solution, the distance between $A_i$ and $M_{ij}$ can be calculated as follows:

$$d_{A_iM_i} = \sqrt{\left(x_{A_i} - x_{M_{ij}}\right)^2 + \left(y_{A_i} - y_{M_{ij}}\right)^2} \tag{4}$$

where $d_{A_iM_i}$ is the collision distance between $A_i$ and $M_{ij}$.

All the collision distance should be calculated at each frame. The minimum value is the actual collision distance between two vehicles when vehicle B is seen as a stationary vehicle.

$$d_{TTC_{AB}} = \min\{d_{A_1M_{11}}, d_{A_1M_{12}}, \cdots, d_{A_4M_{44}}\} \tag{5}$$

where $d_{TTC_{AB}}$ is the minimum collision distance between vehicle A and vehicle B when vehicle B is seen as a stationary vehicle.

It should be noted that all the above steps should be calculated again assuming that vehicle A is a stationary vehicle.

$$d_{TTC_{BA}} = \min\{d_{B_1M_{11}}, d_{B_1M_{12}}, \cdots, d_{B_4M_{44}}\} \tag{6}$$

where $d_{TTC_{BA}}$ is the minimum collision distance between vehicle A and vehicle B when vehicle A is seen as a stationary vehicle.

$$d_{TTC} = \min\{d_{TTC_{AB}}, d_{TTC_{BA}}\} \tag{7}$$

where $d_{TTC}$ is the actual collision distance between vehicle A and vehicle B.

Therefore, TTC can be calculated as the following equation:

$$\text{TTC} = \frac{d_{TTC}}{\sqrt{v_{Ax}^2 + v_{Ay}^2}} \tag{8}$$

In this way, it is not necessary to distinguish the specific collision situation between the two vehicles, so that calculation efficiency can be improved. The threshold of TTC is 3.0s (Hegeman, 2008; Jin et al., 2011; Minderhoud & Bovy, 2001; Qu et al., 2014). The lower the TTC, the higher the level of a collision will be (Mahmud et al., 2017).

## 3 Data description

A field study was conducted at merging sections at the Outer Ring Expressway, Shanghai, China. As Figure 4 shows, the westbound traffic flow merging section is collected and analysed. This site includes one on-ramp and acceleration lane. The videos were recorded with a DJI Mavic 2 Pro on December 26th, 2020. Some requirements should be considered before data collection. For example, cloudy, sufficient lighting conditions and the wind force below 4m per second. Additionally, the videos were recorded at a flight height of 250m. The length of the road segment recorded by videos is 215m.

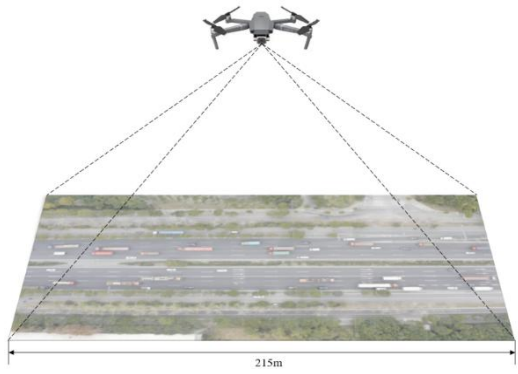
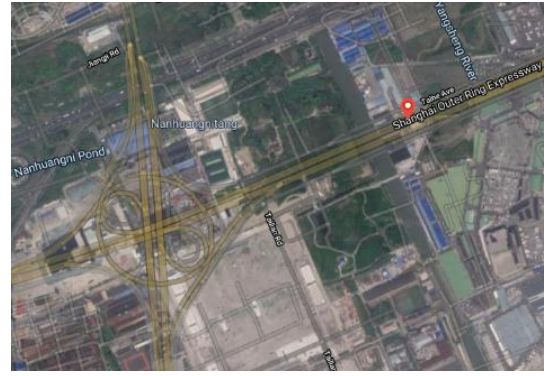

(a) shooting videos by UAV  (b) Location of the field study

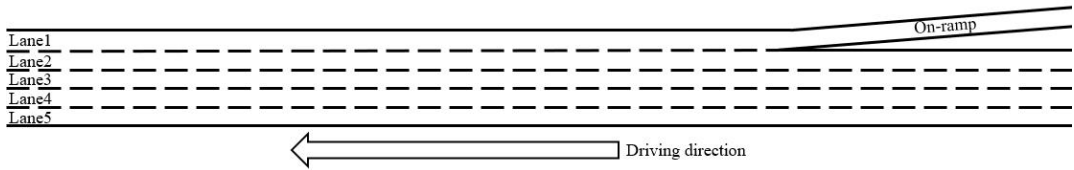

(c) Observation range

Figure 4. Merging sections at the Shanghai Outer Ring Expressway

All videos were recorded in 4K (4096×2160 pixel) resolution at 25 Hz. Total shooting time is 69 min 25s. In this study, vehicles are grouped into two categories, one group is car, and another group is truck. When the length of the vehicle is larger than 6m, the vehicle is classified as a truck. As can been seen from Table 1, the proportion of trucks in our vehicle trajectory data is higher than other existing open-source datasets. Specifically, there are 3858 vehicles in this time period, including 2,512 cars and 1,346 trucks.

Table 1. Comparison of the percentage of trucks between different datasets

| Dataset | Percentage of trucks |
|---|---|
| NGSIM | 3% |
| HighD | 23% |
| Shanghai Outer Ring Expressway | 35% |

# 4 Lane-change conflict analysis

## 4.1 General traffic flow characteristics

The speed distribution is shown in Figure 5. The mean speed of cars is larger than trucks at this merging section. Figure 6 provides the spatial distribution of speed. The spatial distribution of car's speed is similar to that of truck's speed. More specifically, the mean speed of vehicles in the innermost lane is the highest compared with other four lanes. At inner four lanes, the mean speed of vehicles all increase at upstream and then decrease at downstream. Besides, the mean speed at on-ramp is decreased and then gradually increased after passing on-ramp into the outermost lane.

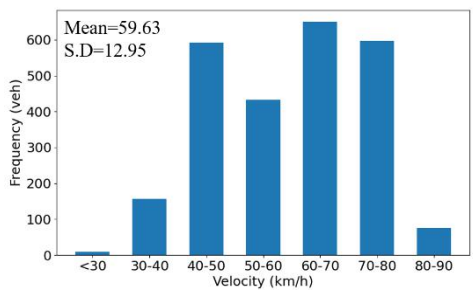
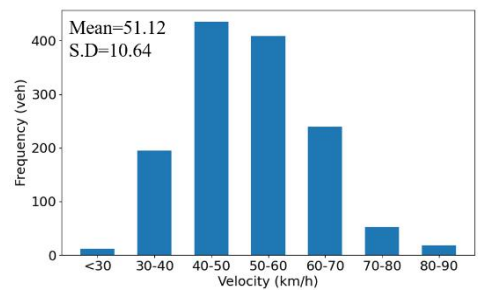

(a) Speed distribution of cars  (b) Speed distribution of trucks

Figure 5. Speed distribution of two vehicle types

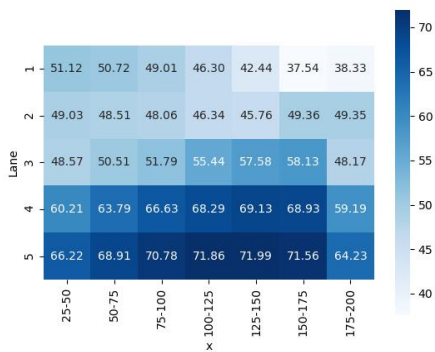
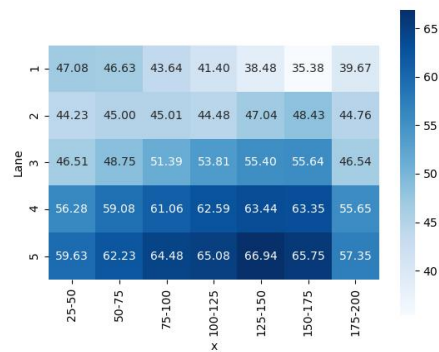

(a) Spatial distribution of car's speed  (b) Spatial distribution of truck's speed

Figure 6. Spatial distribution of speed

## 4.2 Frequency and severity of lane-change conflicts

In this study, TTC is utilized to analyse lane-change conflicts between cars and trucks. There are four different categories of lane-change conflicts. Some lane-change situations are shown in Figure 7.

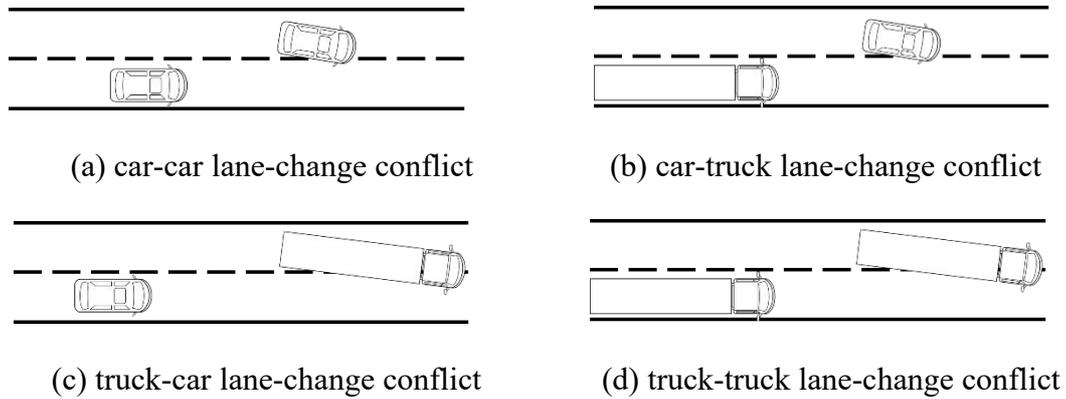

(a) car-car lane-change conflict        (b) car-truck lane-change conflict

(c) truck-car lane-change conflict       (d) truck-truck lane-change conflict

Figure 7. Four categories of lane-change conflicts

Figure 8 shows the number of TTC and the mean value of TTC for different conflict vehicle types at the merging section. Except truck-truck, the conflict number of other three conflict categories are similar. The regularity of TTC mean value is: $TTC_{car\text{-}car} < TTC_{car\text{-}truck} < TTC_{truck\text{-}car} < TTC_{truck\text{-}truck}$. The mean TTC for lane-change conflicts is greater when involving trucks.

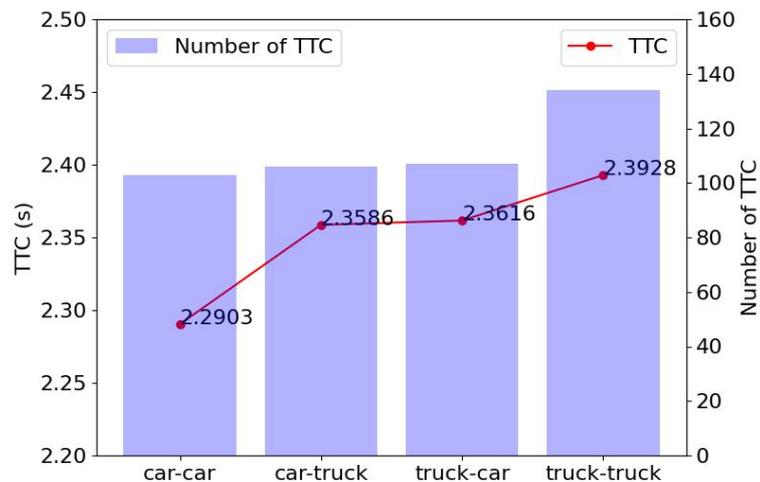

Figure 8. TTC number and mean value for different types of vehicles

When the lead vehicle is a car, consider different types of the lag vehicle. Compared cars with trucks, cars are characterized by higher mobility and flexibility because of their small body volume. When vehicles change lanes, for car-car, the relative distance is smaller or relative speed is higher than other three categories, so mean TTC of car-car conflict is the smallest among four categories. When the lead vehicle is a truck, consider the lag vehicle. If the lead vehicle decelerated, it takes longer time for the truck to decelerate to avoid a lane change collision than a car. Hence, the lag truck driver tends to maintain larger relative distance or smaller relative speed with the lead vehicle than the lag car driver. Considering these factors, the mean TTC of truck-truck is the largest among four categories.

*4.3 Lane-change conflict positions between cars and trucks*

Figure 9 graphically shows the spatial distribution characteristics of lane-change conflicts at merging section. From Figure 9, we can see that few lane-change conflicts occur at upstream inner lanes. Lane-change conflicts frequently occur from on-ramp to acceleration lane, because vehicles must merge into mainline in this location. Besides, vehicle drivers at mainline know that vehicles at on-ramp will merge to mainline so they prefer change lane to inner lanes in order to reduce lane-change conflicts with vehicles from on-ramp. Consequently, in the middle of the segment, lane-change conflict positions gradually move from the upstream outer lanes to the downstream inner lanes. Additionally, almost no lane-change conflicts occur at outermost lane. Lane-change conflicts between four inner lanes of downstream are evenly distributed.

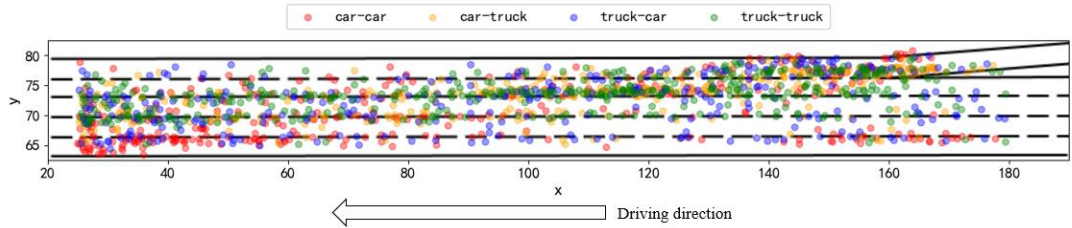

Figure 9. Lane-change conflict positions between cars and trucks

Figure 10 shows the characteristics of lane-change conflict positions for different vehicle types in detail. For car-car lane-change conflicts (Figure 10. (a)), conflicts always occur at the acceleration lane and the downstream inner lanes. Although the number of lanes increases, drivers do not tend to drive at outermost lane. Besides, cars at the upstream tend to change lanes from the outer lane to the inner lane, especially to the innermost lane, because the closer the lane is to the inside, the higher the speed limit. Another important reason is that vehicles at inner lane are less affected by vehicles merging into mainline from on-ramp. Moreover, traffic conflicts between cars become more serious at downstream inner lanes. For car-truck lane-change conflicts (Figure 10. (b)), lane-change conflicts also frequently occur at the acceleration lane, especially some severe conflicts occur at this position. Moreover, car-truck conflicts appear from the upstream outer lane to downstream inner lane. For truck-car lane-change conflicts (Figure 10. (c)), lane-change conflict spatial distribution is similar to car-truck lane-change conflicts. For truck-truck lane-change conflicts (Figure 10. (d)), conflicts rarely occur at inner two lanes. Most truck-truck conflicts occur at outer lanes. Besides, truck-truck conflicts in the two outermost lanes are more serious than the inner lanes.

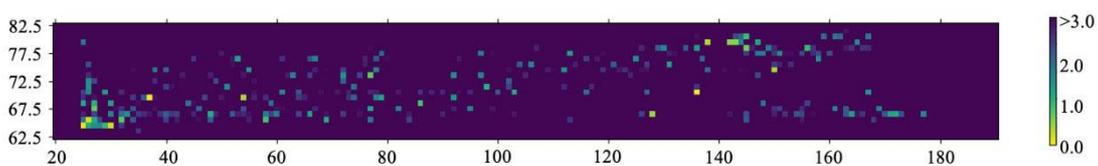

(a) lane-change conflict positions (car-car)

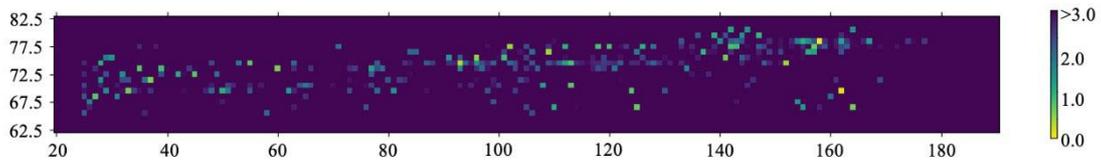

(b) lane-change conflict positions (car-truck)

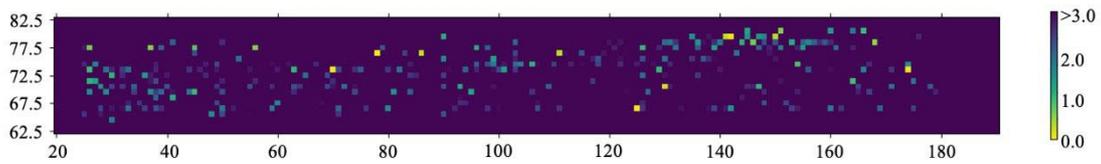

(c) lane-change conflict positions (truck-car)

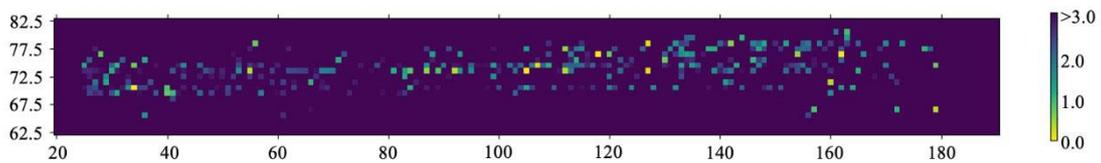

(d) lane-change conflict positions (truck-truck)

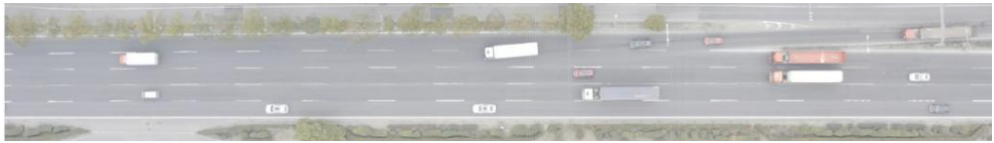

(e) Observed road segment

Figure 10. Lane-change conflict positions for different types of vehicles

**5 Conclusion and findings**

This study comprehensively examined lane-change conflict between different vehicle types at merging sections. A TTC calculation method is utilized to calculate lane-change conflicts. Then, the TTC values between different vehicle types are compared. This study also analyses spatial distribution characteristics of lane-change conflicts.

Results show that due to the significant difference in the operating characteristics of cars and trucks, there are differences in the spatial distribution of traffic conflicts between different vehicle types. The number of truck-truck lane-change conflicts is the largest, but these this type of conflicts is less severe compared with other type of conflicts. Although the number of car-car lane-change conflicts is smallest, the severity is the highest. For the spatial distribution characteristics of lane-change conflicts, conflicts always occur from on-ramp to acceleration lane. It means this position have a high risk of accidents. Specifically, conflict positions of vehicles vary from vehicle types to vehicle types. Car-car conflicts always occurs at downstream inner lanes. Both of car-truck conflicts and truck-car conflicts appear from the upstream outer lane to downstream inner lane. Truck-truck conflicts always occur at outer lane. Based on the above results, some useful strategies can be proposed to reduce lane-change conflicts at merging section. For example, since serious car-car conflicts occur in the downstream inner lane, an information board can be installed on the gantry of the inner lane to remind cars to maintain safe distance. Transportation management agencies are suggested to implement traffic management at on-ramp freeway merging section, such as change dotted line to solid line at the beginning of acceleration lane so that vehicles can not change lanes at this position. Also, for connected vehicles, results could be used to suggest drivers especially car drivers maintaining safe distance and choose a suitable lane to drive. This study is not without limitations. Only one road segment was selected to analyse, and other road segments can be chosen to study to verify the findings in the future.